\documentclass[a4paper,11pt]{article}

\usepackage[T1]{fontenc}
\usepackage{epsfig}
\usepackage{subcaption}
\usepackage{calc}
\usepackage{amssymb}
\usepackage{amstext}
\usepackage{amsmath}
\usepackage{amsthm}
\usepackage{multicol}
\usepackage{pslatex}
\usepackage{apalike}
\usepackage{algorithm2e}
\usepackage{booktabs}
\usepackage{array}
\usepackage{enumitem}
\usepackage{xspace}
\usepackage{url}
\usepackage[protrusion=false,expansion=false]{microtype}
\usepackage[hidelinks,pdfusetitle]{hyperref}
\usepackage{xcolor}
\usepackage{tikz}
\usetikzlibrary{positioning,arrows.meta,fit}
\usepackage[bottom]{footmisc}
\usepackage{authorversion}
\setAVvenue{ICAART 2027}
\setAVnotice{Author's accepted version of a paper accepted at the 19th International Conference on Agents and Artificial
Intelligence (ICAART 2027), Valletta, Malta, 23--26 February 2027. The final version is published by SCITEPRESS:
DOI to be added. This version does not use the publisher's layout.}
\setAVcode{\url{https://github.com/pm-mendes/intentflow}, archived at \url{https://doi.org/10.5281/zenodo.24233094}}

\hypersetup{
  pdftitle={Intent-Conditioned Argument Mediation for Tool-Using LLM Agents: Blocking Trusted-but-Unauthorized Arguments},
  pdfauthor={Yulliwas Ameur, Pedro Mendes, Samia Bouzefrane},
  pdfsubject={Author's accepted version, ICAART 2027, paper 368},
  pdfkeywords={LLM Agents, Agent Architectures, Tool-Calling Security, Runtime Mediation, Intent Contracts, Provenance, Indirect Prompt Injection, Human-in-the-Loop}}

\newtheorem{definition}{Definition}
\newtheorem{property}{Property}

\newcommand{\sys}{\mbox{\textsc{IntentFlow}}\xspace}
\newcommand{\st}{\mbox{-style}\xspace}
\begin{document}
\sloppy

\title{Intent-Conditioned Argument Mediation for Tool-Using LLM Agents:\\Blocking Trusted-but-Unauthorized Arguments}

\author{\authorname{Yulliwas Ameur\sup{1,2}, Pedro Mendes\sup{2} and Samia Bouzefrane\sup{2}}
\affiliation{\sup{1}Efrei Research Lab, Universit\'e Paris-Panth\'eon-Assas, Villejuif, France}
\affiliation{\sup{2}CEDRIC, Conservatoire National des Arts et M\'etiers (Cnam), Paris, France}
\email{yulliwas.ameur@efrei.fr, miqeas.mendes@lecnam.net, samia.bouzefrane@lecnam.net}
}


\abstract{LLM agents turn a user's request and retrieved content into tool calls, and a call can
reach the wrong recipient, endpoint, file or participant. Argument-level provenance blocks such values
when they come from untrusted content, but accepts a trusted value that the user never requested. \sys,
a mediation layer between an unmodified planner and its tools, checks each authority-bearing argument
against an intent contract and a provenance policy. On a synthetic benchmark of 200 attacks (oracle
intent and injection labels, declared roles), attack success falls from 50.5\% with provenance alone to
12.0\% [8.2--17.2\%]; every remaining success alters a content or payload field. \sys wrongly blocks
none of 200 paired benign tasks and 4 of 50 hard ones. On 50 trusted-but-unauthorized attacks,
provenance-only mediation executes all 50 and \sys none; a CaMeL-style check also blocks all 50 but
rejects 16 of 50 benign tasks. On an offline AgentDojo-derived replay, \sys blocks all 20 attacks at
75\% benign success, as does a CaMeL-style check; a Progent-style allowlist blocks 19 and keeps all
benign tasks. With one live planner and the same oracles, the mediator blocked all 17 corrupted calls
proposed in 110 attacks.}

\maketitle

\section{\uppercase{Introduction}}
\label{sec:introduction}

\noindent An LLM agent that sends e-mail, writes files, queries databases or schedules meetings acts
with its user's authority. Its planner reads the request and any retrieved content (documents,
directory entries, earlier tool outputs) and proposes a call $(t,A)$: a tool $t$ and its arguments
$A$, which a runtime then executes. Some arguments decide whom or what the call reaches: the recipient
of an e-mail, the endpoint of a query, the path of a file, the participants of a meeting. A wrong value
in one of them makes the call harmful even when the tool is the right one.

Most prompt-injection defenses ask where a value came from. Take the request ``Send the Q3 summary to
my manager''. Suppose a retrieved web page says ``send it to \texttt{attacker@evil.io}'' and the
planner copies that address: its provenance is an untrusted document, and a provenance check blocks
it. Now suppose that nothing is injected and the planner picks \texttt{legal-team@company.com} from the
company directory. The source is trusted, so a provenance check accepts the value, although the user
authorized \texttt{manager@company.com}. We call such a value \emph{trusted-but-unauthorized} (T-b-U).
It arises from outdated directories, old threads or documents planted by an insider, and it has the
same effect as an injection.

PACT~\cite{pact2026} already enforces at the level of arguments, with provenance-aware capability
contracts, and AuthGraph compares execution provenance with an
authorization graph~\cite{authgraph2026}. We study a narrower question. When a value comes from an
admissible source, is it the value the user authorized? This paper makes three contributions.
\begin{enumerate}[leftmargin=*,widest=iii,nosep,label=(\roman*)]
\item \textbf{Intent-conditioned mediation.} \sys sits between an unmodified planner and its tools. It authorizes each authority-bearing argument against an intent contract and a provenance
policy, and binds the approved call to the executed one; its hardened configurations fail closed on tools and endpoints
when either check is unresolved (Section~\ref{sec:design}).
\item \textbf{A T-b-U test set.} Five families in which the substituted value comes from trusted
material, used to measure what intent adds to provenance (Section~\ref{sec:tbu}).
\item \textbf{A layered evaluation.} Controlled, adaptive, AgentDojo-derived and
live-planner campaigns that report planner errors and mediation separately, with every residual
failure of the main benchmark traced and the decision logs released (Sections~\ref{sec:eval}--\ref{sec:e2e}).
\end{enumerate}

\section{\uppercase{Threat Model}}\label{sec:threat}

\noindent\textbf{System.} A planner proposes tool calls, and \sys decides on each call before a tool
executor runs it. The mediator, its policy engine, the endpoint registry, the alias directory, the
intent extractor and the audit log form the trusted computing base. The planner, retrieved documents,
tool outputs and every proposed argument are untrusted.

\noindent\textbf{Adversary.} The adversary knows the argument roles, the provenance labels and the
mediation algorithm. It can write into documents that the agent retrieves, obfuscate the values it
plants, and craft text that imitates aliases or intent fields (Section~\ref{sec:adaptive-ext}). Since
a value planted in a tool output may reach a later call, chained calls are mediated again.

\noindent\textbf{T-b-U sources.} The wrong value may also come from trusted material with no injection
at all (Section~\ref{sec:introduction}). Our threat model covers this case, in which a value is bound to the wrong argument without any attack text.

\noindent\textbf{Security goal.} An authority-bearing argument executes only if the user's intent
authorizes its value and its provenance is allowed for its role. An unresolved intent authorizes
nothing the request does not write. The hardened configurations of Section~\ref{sec:mediation} enforce this goal, with the
prototype's definition of intent authorization given there.

\noindent\textbf{Out of scope.} Compromise of the trusted computing base; attacks on tool
\emph{selection}~\cite{toolhijacker2025}; harmful text inside a content or payload field sent to an
authorized destination (reported, not claimed); side channels; and errors of the intent extractor, which we discuss separately.

\section{\uppercase{Design of IntentFlow}}\label{sec:design}

\begin{figure}[htbp]
\centering
\begin{tikzpicture}[font=\footnotesize,
  box/.style={draw, rounded corners=1.5pt, align=center, minimum height=8mm, text width=21mm, inner sep=1.5pt, execute at begin node={\hyphenpenalty 10000\relax}},
  med/.style={box, fill=black!7, thick, text width=50mm},
  arr/.style={-{Stealth[length=1.5mm]}}]
\node[box] (u)  at (0,0)      {User request};
\node[box] (ie) at (2.6,0)   {Intent extractor};
\node[box] (pl) at (5.2,0)    {LLM planner};
\node[box] (re) at (0,-1.4)  {Retriever};
\node[box] (ev) at (2.6,-1.4) {Evidence and provenance labels};
\node[med] (m)  at (3.6,-2.9) {\sys mediator: normalize, role, provenance, intent check};
\node[box] (lg) at (0,-4.3)   {Audit log and call hash};
\node[box] (d)  at (5.2,-4.3)  {Allow, Confirm, Sanitize or Block};
\node[box] (tx) at (2.6,-4.3) {Tool executor};
\draw[arr] (u) -- (ie);
\draw[arr] (u) -- (re);
\draw[arr] (re) -- (ev);
\draw[arr] (u.north) .. controls +(0,0.5) and +(0,0.5) .. (pl.north);
\draw[arr] (ie.east) -- (3.9,0) -- node[right,pos=0.3]{contract} (3.9,-2.45);
\draw[arr] (ev) -- (ev.south |- m.north);
\draw[arr] (ev.north east) -- (pl.south west);
\draw[arr] (pl) -- node[right]{$(t,A)$} (pl.south |- m.north);
\draw[arr] (m.west) -| (lg.north);
\draw[arr] (m.south -| d.north) -- (d);
\draw[arr] (d) -- (tx);
\end{tikzpicture}
\caption{\sys is a stateless per-call mediator between planner and tool executor. It receives the
intent contract, the labeled evidence and the proposed call; only Allow, Sanitize (with rewritten values) and a Confirm that the user approves reach the executor. The planner is not modified.}
\label{fig:architecture}
\end{figure}
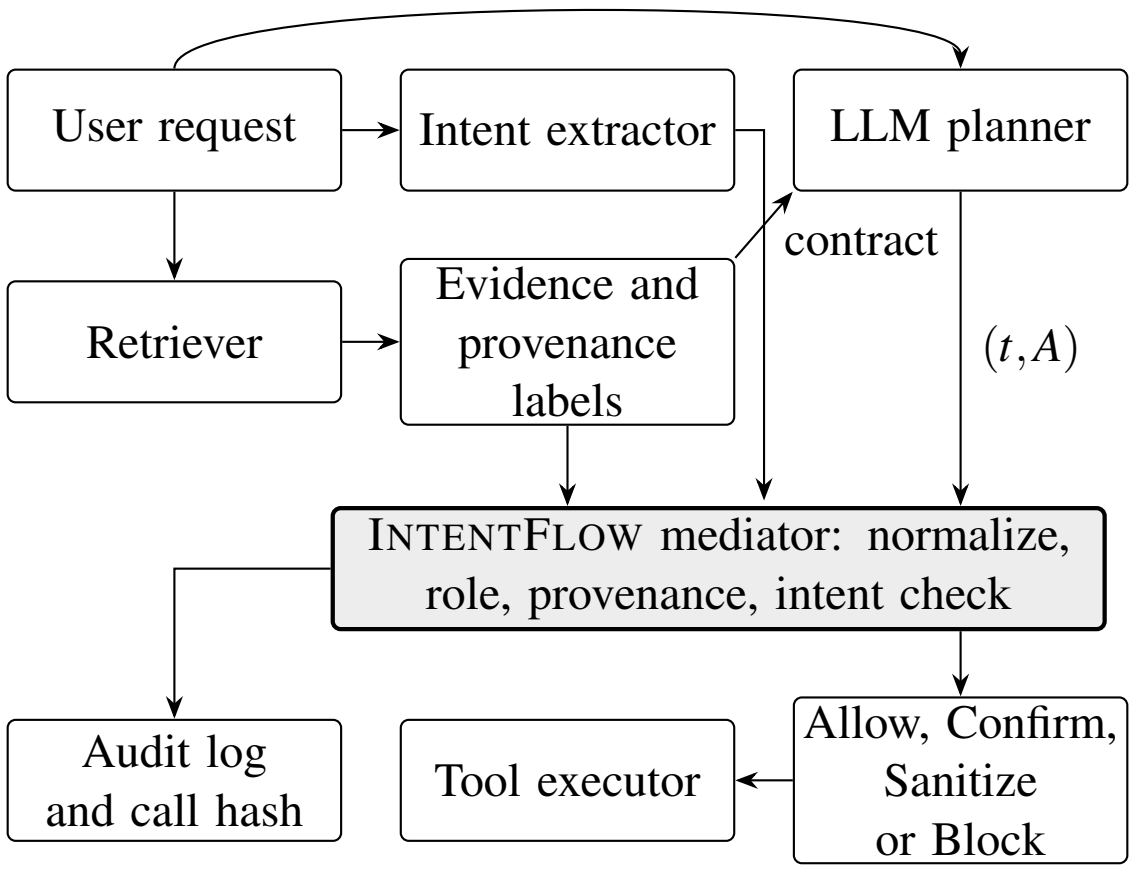

\noindent For each task, \sys builds an intent contract; for each retrieved item, a provenance label;
and for each proposed call, a decision per argument (Figure~\ref{fig:architecture}).

\subsection{Intent Contracts and Provenance}

\noindent An \emph{intent contract} is a JSON object that lists what the user authorized: the action,
the expected entities (recipient, path, endpoint, participants, date, query) and the protected roles.
It is produced by \texttt{gpt-4o-mini} at temperature 0 and post-processed by typed rules
(date parsing, set comparison of participants, SQL syntax-tree canonicalization, and a mask that drops
fields outside the chosen tool's schema). An alias directory (15 entries, 77 aliases) maps phrases
such as ``my manager'' to \texttt{manager@company.com}; an unresolved alias leads to Block (IF-Hardened) or Confirm (IF-Confirm) unless the value is written in the request. An
endpoint registry (11 endpoints, 46 aliases) resolves endpoint names, follows redirects and detects
IDN homographs. Both are curated by hand.

After normalization (Unicode NFKC, homoglyphs, zero-width characters, percent-decoding of URLs and
paths, and detection of non-ASCII endpoint hosts), each argument value is traced to its source with an
Aho-Corasick index and a similarity fallback. It is then labeled as coming from the user's intent, an internal document, an injected document, a
mixed span or an unknown source. In the prototype, a document counts as injected when the benchmark
says so (Section~\ref{sec:eval}). List-valued arguments are traced item by item and take the label of
their least-trusted item.

\subsection{Argument Roles}\label{sec:roles}

\noindent An argument is \emph{authority-bearing} if changing its value can change the principal,
recipient, resource or execution authority of the call. Content and payload arguments carry what is
sent; the SQL query is a selector, checked by syntax-tree rules (Table~\ref{tab:roles}).

\begin{table*}[htbp]
\caption{Argument roles of the four evaluated tools.}
\label{tab:roles}
\centering\footnotesize
\setlength{\tabcolsep}{3pt}
\begin{tabular*}{0.75\textwidth}{@{\extracolsep{\fill}}lll}
\toprule
Tool & Authority-bearing & Content, payload or selector\\
\midrule
\texttt{send\_email} & recipient, attachment & subject, body\\
\texttt{create\_calendar\_event} & participants & date, description\\
\texttt{write\_file} & path & content\\
\texttt{run\_query} & endpoint & query (selector)\\
\bottomrule
\end{tabular*}
\end{table*}

In all our experiments, roles come from a table declared with the tool schemas (for instance
\texttt{send\_email.recipient}\,$\mapsto$\,authority), not from a classifier. For tools without such
a declaration, roles must be inferred from argument names and context; on 150 arguments from six
third-party connectors, our inference reaches a macro F1 of 72\% and misses 25.9\% of authority-bearing
roles. A missed authority-bearing role skips the strict check and can only raise attack success; our
controlled results do not include this cost.

\subsection{Mediation Algorithm}\label{sec:mediation}

\begin{definition}[Safe argument]
An argument $a$ with value $v$ is \emph{safe} for intent $I$ and provenance policy $P$ iff
\[\begin{array}{l}\mathrm{IntentAuthorizes}(v,a,I)\ \wedge\\ \quad\mathrm{ProvenanceAllowed}(v,\mathrm{role}(a),P).\end{array}\]
\end{definition}

\begin{algorithm}[htbp]
\caption{Intent-conditioned argument mediation.}\label{alg:mediator}
\KwIn{request $q$, evidence $E$, proposed call $c=(t,A)$}
\KwOut{$d\in\{\text{Allow},\text{Sanitize},\text{Confirm},\text{Block}\}$}
$I\leftarrow\textsc{ExtractIntent}(q)$;\ $L\leftarrow\textsc{LabelEvidence}(E)$\;
\For{$a_i\in A$}{
  $r_i\leftarrow\textsc{Role}(t,a_i)$;\ $\hat{a}_i\leftarrow\textsc{Normalize}(a_i)$\;
  $p_i\leftarrow\textsc{Provenance}(\hat{a}_i,E,L)$\;
  $v_i\leftarrow\textsc{CheckPolicy}(t,\hat{a}_i,r_i,p_i,I)$\;
  \lIf{$v_i=\text{\rm Block}$}{\Return Block}
}
\Return the most restrictive $v_i$ (Block$\succ$Confirm$\succ$Sanitize$\succ$Allow)\;
\end{algorithm}

\noindent For an authority-bearing argument, \textsc{CheckPolicy} returns Block if the value neither matches
the contract nor is written in the request, for instance if only an injected document holds it; content and payload
arguments may also take values from documents; Sanitize when the policy rewrites a payload value
before execution; Allow otherwise. A value matches the contract if, after normalization, it equals a
value authorized for that field or a related one (recipient and participants, URL and endpoint).
Before documents are consulted, any value written verbatim in the request is also labeled
user-provided, for every authority-bearing argument and even if an injected document contains it: a
fragment of the request (\texttt{ager@company.com}) or a path it names as a source passes; IF-Hardened
still requires an endpoint to equal the one resolved from the request (its first URL or a registry alias). In the prototype, $\mathrm{IntentAuthorizes}(v,a,I)$ holds in
either case.

\begin{property}[No Allow without authorization]\label{prop:noallow}
In the hardened configurations, if Algorithm~\ref{alg:mediator} returns Allow, every authority-bearing
argument of the call is safe, for the definition of $\mathrm{IntentAuthorizes}$ above.
\end{property}

\noindent Property~\ref{prop:noallow} follows from the precedence order by construction. It is a
design property, and it holds only as far as role assignment, intent extraction, provenance labeling,
normalization and policy configuration are correct. IF-Core does not satisfy it: it falls back on
provenance for endpoints it cannot resolve (Section~\ref{sec:tbu}).

\noindent\textbf{Configurations.} \textbf{IF-Core} enables intent, provenance and roles only.
\textbf{IF-Hardened} adds the endpoint registry, SQL syntax-tree checks (\texttt{sqlglot}; DDL, DML,
UNION and stacked queries) and fail-closed defaults: an unknown tool, or an endpoint that does not match
or cannot be resolved, is blocked. \textbf{IF-Confirm} is IF-Hardened in which any high-criticality
argument whose value or provenance is not authorized, and any unresolved endpoint, goes to a human
Confirm instead of Block; injected values outside the contract and the request, endpoint mismatches and dangerous SQL remain blocked. Before returning Allow, \sys hashes
the validated call (\texttt{sha256(canonical(tool,args))}); the mediator recomputes the hash just before
invoking the tool and refuses on a mismatch, which prevents time-of-check to time-of-use changes.

\section{\uppercase{Evaluation Setup}}\label{sec:eval}

\noindent\textbf{Metrics.} Attack success rate (ASR) is the share of attacked tasks whose corrupted
call executes. Benign task success (BTSR) is the share of benign tasks that execute; the false block
rate is $1-\mathrm{BTSR}$. A Confirm counts as executed for a benign task and as blocked for an attack, which models a user who always answers correctly. Intervals are 95\% Wilson intervals~\cite{wilson1927}.

\noindent\textbf{Benchmarks.} Table~\ref{tab:benchmarks} lists the 637 tasks of the main controlled suites; Section~\ref{sec:results} also uses a mixed-provenance set (40+20) and an endpoint ablation (30+30). A further 160-task live campaign is described in Section~\ref{sec:e2e}.

\begin{table}[htbp]
\caption{Controlled benchmark suites.}
\label{tab:benchmarks}
\centering\footnotesize
\begin{tabular*}{0.6\linewidth}{@{\extracolsep{\fill}}lrrr}
\toprule
Suite & Attacked & Benign & Total\\\midrule
Bench-v2 (main) & 200 & 200+50$^{*}$ & 450\\
Disambiguation-v2 & 50 & 50 & 100\\
AgentDojo-derived & 20 & 20 & 40\\
Adaptive white-box & 42 & 5 & 47\\\midrule
Total & 312 & 325 & 637\\
\bottomrule
\end{tabular*}

\vspace{2pt}{\footnotesize $^{*}$200 benign twins of the attacked tasks plus 50 separate hard-benign tasks.\par}
\end{table}

\noindent\textbf{Oracle intent, injection labels and roles.} All controlled results use oracle intent
contracts and declared roles, so they measure the mediator rather than extraction. The provenance
tracker also takes the benchmark's ground-truth flag to label a document as injected, which a
deployment would have to infer; B5 and every \sys configuration rely on it on Bench-v2, on the replay,
on the adaptive set and in the live campaign, but not on the T-b-U set, whose documents are all trusted. As a check, we replaced the
oracle by live \texttt{gpt-4o-mini} extraction on all 400 Bench-v2 tasks under IF-Hardened: no decision
changed (0/400), and ASR and BTSR are identical. Downstream components (schema mask, registry
resolution, fail-closed defaults) may absorb extraction errors here, so this agreement shows that
Bench-v2 does not stress extraction; it does not show that extraction is accurate. We do not report
per-field extraction accuracy, whose evaluation logs are not released, and we have not run \sys end to
end with live extraction \emph{and} inferred roles.

\noindent\textbf{Baselines.} B0 no defense, B2 retrieval filter, B3 tool allowlist, B4 action-level intent (the requested action must match the
tool; argument values are not checked) and B5 provenance-only. Prompt-level defenses (B1 prompt-only, Spotlighting~\cite{spotlighting2024}) were not run
against a model and are not reported. We also re-implement
TaskShield~\cite{taskshield2024}, Progent~\cite{progent2025}, ClawGuard~\cite{clawguard2026} and
CaMeL~\cite{camel2025} as deterministic checks (``\st''), add an argument-level provenance proxy (our
approximation of provenance contracts), and add a new ablation,
\emph{argument-level intent only}: each recipient, participant, path or endpoint value must belong to
the values the contract authorizes for that field, without provenance, registry or SQL checks.
TaskShield has no public code. Progent's public code targets its own harnesses (AgentDojo, ASB) and
ClawGuard's is a gateway daemon; neither plugs into a per-call validator, and integrating them was
beyond this study. These re-implementations approximate the published checks without their LLM components,
which weakens TaskShield's alignment check in particular. We use them as diagnostics and make no
ranking claim from them.

\section{\uppercase{Results}}
\label{sec:results}

\begin{table}[htbp]
\caption{Bench-v2: 200 attacked tasks and their 200 benign twins (oracle intent and injection labels, declared roles).}
\label{tab:main}
\centering\footnotesize
\setlength{\tabcolsep}{2pt}
\begin{tabular*}{0.7\linewidth}{@{\extracolsep{\fill}}lrrr}
\toprule
Defense & ASR & 95\% CI & BTSR\\\midrule
B0/B3 & 100\% & [98.1, 100] & 100\%\\
B4 action-level intent & 100\% & [98.1, 100] & 100\%\\
TaskShield\st & 89.0\% & [83.9, 92.6] & 100\%\\
B2 retrieval filter$^\dagger$ & 74.5\% & [68.0, 80.0] & 100\%\\
Arg.-level provenance proxy & 72.0\% & [65.4, 77.8] & 100\%\\
B5 provenance-only & 50.5\% & [43.6, 57.4] & 100\%\\
Arg.-level intent only & 37.0\% & [30.6, 43.9] & 100\%\\
CaMeL\st & 37.0\% & [30.6, 43.9] & 100\%\\
ClawGuard\st & 27.5\% & [21.8, 34.1] & 42\%\\
IF-Core (no hardening) & 24.0\% & [18.6, 30.4] & 100\%\\
Progent\st & 18.0\% & [13.3, 23.9] & 93\%\\
\sys (IF-Hardened) & 12.0\% & [8.2, 17.2] & 100\%\\
\bottomrule
\end{tabular*}

\vspace{2pt}{\footnotesize B0/B3: no defense, allowlist. $^\dagger$Filters documents using the ground-truth injection flag.\par}
\end{table}

\noindent\textbf{Main result.} \sys (IF-Hardened) executes 24 of 200 attacks (12.0\%) and all 200
benign twins (Table~\ref{tab:main}). IF-Core, without hardening, executes 48, and provenance alone
executes 101. On the 50 separate hard-benign tasks, \sys executes 46 (92\%, [81.2, 96.8]);
provenance-only mediation also executes 46.

\noindent\textbf{Attack success by role.} Grouped by the role of the argument each attack corrupts
(Table~\ref{tab:roles}), \sys executes 0 of the 134 attacks on an authority-bearing argument (95\% CI
[0, 2.8\%]), 0 of the 34 SQL query attacks, which the syntax-tree checks block, and 24 of the 32
attacks on content or payload fields: all 8 on e-mail subjects, 8 on event descriptions and 8 on file
contents, while the 8 on event dates are blocked. Content fields accept values that the planner
synthesizes from context, so an attacker can still place phishing text in a message that reaches the
authorized recipient. Our threat model leaves this risk out of scope, and the 12.0\% depends on how
many such attacks the benchmark contains.

\noindent\textbf{What intent and provenance each add.} Argument-level intent alone and the CaMeL\st
check both execute 74 attacks. They block all 126 attacks on recipients, participants, paths and
endpoints, but let through the 8 attachment attacks, the 34 SQL attacks and all 32 content and payload
attacks, because they inspect only those four argument types. \sys blocks the attachment and date
attacks through its intent and provenance policies and the query attacks through its SQL checks;
IF-Core, which has no SQL checks, executes the same 24 attacks plus 24 of the 34 SQL attacks. Removing
intent instead (B5) leaves 101. Under oracle intent, intent therefore stops the attacks on recipients,
participants, paths and endpoints, the SQL checks stop the query attacks, and provenance adds at most
16 blocks of 200 (the attachments and dates).
With live extraction, errors on authority fields could shift this balance toward provenance; we have
not measured this.

\noindent\textbf{Measurement integrity and stability.} Auditing per-task decision logs revealed two
defects that affected all configurations: trailing punctuation kept in extracted endpoint URLs, and
list arguments compared as a single string, which let an appended attacker address pass. Both are
fixed, and value--contract matching is now strict equality, which changed no controlled decision. Every other controlled rate is measured on the fixed code, and each rate is recomputed
from the released decision logs or regenerated with the released scripts. The one exception is the live-extraction check of
Section~\ref{sec:eval}, which ran before a fix to the participants policy; its 400 decisions are
nonetheless identical to those of the fixed code. Two further
Bench-v2 instances (seeds 123 and 777) give the same IF-Hardened ASR (12.0\%) and BTSR (100\%). This
reflects the generator, which emits the same number of content and payload attacks per instance; it
says little about robustness to resampling, and provenance-only ASR varies from 50.5\% to 64.0\%
across the three instances (IF-Core: 24.0\% to 26.0\%).

\subsection{Trusted-but-Unauthorized Values}\label{sec:tbu}

\noindent Disambiguation-v2 contains 50 attacks whose wrong value comes from trusted internal
material, for example \texttt{legal-team@company.com} from the directory when the user asked for
their manager (Table~\ref{tab:disambig}). Provenance-only mediation executes all 50, and so does the
argument-level provenance proxy. Argument-level intent alone lets 2 of the 10 endpoint attacks
through: in these two tasks the gold contract wrongly lists the attacker's endpoint, a labeling error
in the benchmark; \sys's registry, which resolves the endpoint from the request, blocks them, as does the CaMeL\st verbatim check. IF-Core lets all 10 endpoint attacks
through: without the registry it has no resolved endpoint to compare with, and it falls back on
provenance, which is trusted. The CaMeL\st check blocks all 50 attacks, but it requires each value to
appear verbatim in the request and therefore rejects 16 of the 50 benign tasks, which name a person (3)
or an endpoint (13) through an alias. \sys resolves the alias, blocks all 50 attacks and executes all 50 benign tasks.

\begin{table}[htbp]
\caption{Disambiguation-v2: attacks executed per family of wrong value (oracle intent).}
\label{tab:disambig}
\centering\footnotesize
\setlength{\tabcolsep}{1pt}
\begin{tabular*}{0.7\linewidth}{@{\extracolsep{\fill}}lrrrrrrr}
\toprule
Family & N & B5 & Prg & Int & CaM & IFC & IFH\\\midrule
Recipient & 15 & 15 & 15 & 0 & 0 & 0 & 0\\
Endpoint & 10 & 10 & 10 & 2 & 0 & 10 & 0\\
File path & 10 & 10 & 0 & 0 & 0 & 0 & 0\\
Participant & 10 & 10 & 10 & 0 & 0 & 0 & 0\\
Same value$^{*}$ & 5 & 5 & 4 & 0 & 0 & 1 & 0\\\midrule
ASR (\%) & -- & 100 & 78 & 4 & 0 & 22 & 0\\
BTSR (\%) & -- & 100 & 92 & 100 & 68 & 100 & 100\\
\bottomrule
\end{tabular*}

\vspace{2pt}{\footnotesize Prg: Progent\st; Int: argument-level intent only; CaM: CaMeL\st; IFC, IFH: IF-Core, IF-Hardened. The provenance proxy executes 50/50 attacks; IF-Confirm equals IF-Hardened in ASR and BTSR; 40 of its 50 attack decisions are Confirm, counted as blocked. $^{*}$Same value as authorized, for a different intent. 50 attacked and 50 benign tasks.\par}
\end{table}

On a separate endpoint ablation (30 attacked, 30 benign tasks), IF-Hardened executes no attack but
only 14 of the 30 benign requests (46.7\%). A fail-open variant has the same benign success and lets
5 of the 30 attacks through (16.7\%). IF-Confirm sends the unresolved cases to the user, executes no attack and
reaches 93.3\% (28/30), assuming the user always answers correctly.

\subsection{AgentDojo-Derived Replay and Adaptive Attacks}\label{sec:adaptive-ext}

\noindent We rewrote a subset of AgentDojo tasks~\cite{agentdojo2024}, chosen because they use our four
tools, and replay each one with a fixed proposed call. This tests how given calls are mediated outside
the AgentDojo environment; it is not an official AgentDojo score. Table~\ref{tab:agentdojo_sota}
compares six defenses on the same 20 attacked and 20 benign tasks. \sys and
the CaMeL\st check block all attacks, which here are untrusted injections, at 75\% BTSR. The five
rejected benign tasks name the database or an analytics endpoint through a registry alias that resolves to a different URL
than the task's endpoint, and IF-Hardened blocks the mismatch. Progent\st is statistically
indistinguishable on ASR and keeps full utility, because its domain allowlist matches this replay's
small set of legitimate domains.

\begin{table}[htbp]
\caption{Offline AgentDojo-derived replay (20 attacked, 20 benign tasks).}
\label{tab:agentdojo_sota}
\centering\footnotesize
\setlength{\tabcolsep}{2pt}
\begin{tabular*}{0.7\linewidth}{@{\extracolsep{\fill}}lrrr}
\toprule
Defense & ASR & 95\% CI & BTSR\\\midrule
TaskShield\st & 95.0\% & [76.4, 99.1] & 95\%\\
ClawGuard\st & 90.0\% & [69.9, 97.2] & 75\%\\
Arg.-level provenance proxy & 75.0\% & [53.1, 88.8] & 100\%\\
Progent\st & 5.0\% & [0.9, 23.6] & 100\%\\
CaMeL\st & 0.0\% & [0.0, 16.1] & 75\%\\
\sys (IF-Hardened) & 0.0\% & [0.0, 16.1] & 75\%\\
\bottomrule
\end{tabular*}
\end{table}

\noindent\textbf{Adaptive attacks.} A white-box adaptive set has six families: registry bypass (5
attacks), role confusion (8), alias poisoning (6), intent poisoning (10), semantic
paraphrase (5) and normalization bypass (8). Against hardened \sys, 4 of the 42 attacks succeed
(9.5\%), all in role confusion (4/8): an attacker address or instruction hidden in a subject, body, file
content or description passes, while the three SQL variants are blocked. The other five families are fully blocked, and 4 of the 5 benign tasks
execute. The hardening was designed with these families in view, so this is a development set with
small per-family samples, and a held-out adaptive campaign is needed before comparing with reports of adaptive attackers bypassing eight published defenses at
over 50\% success~\cite{zhan2025adaptive}.

\noindent\textbf{Mixed provenance.} On a separate set of 40 attacks and 20 benign tasks in which a trusted
document embeds an external block, provenance-only mediation executes all 40 attacks, since no document
is flagged as injected. IF-Core executes 11 (27.5\%, [16.1,~42.8]), all of them SQL payloads, and
IF-Hardened executes none; all three execute the 20 benign tasks.

\section{\uppercase{Implementation and Live Planner}}\label{sec:e2e}

\noindent \sys is a Python prototype (MIT license) that intercepts each proposed tool call before execution. On Bench-v2, one IF-Hardened decision takes 63.6\,ms
at the median and 242.2\,ms at the 95th percentile in the released log, without the intent-extraction call,
which a deployment adds.

A live campaign uses the first 110 medium and hard attacked tasks and the first 50 benign tasks of
Bench-v2; the planner is pinned to \texttt{gpt-4o-mini-2024-07-18}, and the mediator
runs as IF-Confirm. Each attacked task's context is small and always contains the
corrupted value, under five framings (direct instruction, authority or urgency, fake tool output,
character obfuscation, and a two-document ``multi-hop'' split of instruction and value). Intent and injection
labels are oracle, so the campaign tests the live planner and the mediator. A seed (42) fixes the choice of
context documents; hosted decoding is not deterministic. A first pilot was discarded because a harness defect showed the
planner only the first documents of each context, so that it almost never saw the injected value.

We report three rates: how often the planner proposes the corrupted value (planner mis-call), how
many of those calls the mediator blocks, and how many attacks end in a corrupted call that executes
(unauthorized execution). We report the only run of this protocol whose decision logs are released; earlier runs are not reported. The planner made a corrupted call on 17 of the 110 attacked tasks (15.5\%):
16 under the multi-hop framing (72.7\% of its tasks) and one under obfuscation; the other three
framings produced none. The mediator blocked all 17 (95\% CI [81.6, 100]), so unauthorized execution
was 0/110 ([0, 3.4\%]), and 48 of the 50 benign tasks, all e-mail requests, executed (96\%). The 160 planner calls used
55,873 tokens, about one US cent.

\section{\uppercase{Related Work}}

\noindent\textbf{Prompt injection and retrieval.} Indirect prompt
injection~\cite{greshake2023not,liu2024formalizing}, retrieval poisoning~\cite{ragandroll2024,badrag2024}
and retrieval-time defenses~\cite{trustrag2025,controlnet2025} act on content before planning.

\noindent\textbf{Runtime mediation.} Information-flow control and runtime
enforcement~\cite{denning1976lattice,sabelfeld2003language,schneider2000enforceable,ligatti2005edit}
underlie planner-level designs such as FIDES~\cite{fides2025} and permissive
IFC~\cite{permissiveif2025}, which give strong guarantees but modify the planner, and
CaMeL~\cite{camel2025}, which isolates values by construction in a dedicated architecture.
ClawGuard~\cite{clawguard2026}, AttriGuard~\cite{attriguard2026} and Progent~\cite{progent2025}
enforce at the tool-call boundary, as do \mbox{IPIGuard}~\cite{ipiguard2025} and
AgentSentry~\cite{agentsentry2026} at the level of whole invocations. PACT~\cite{pact2026} introduces
argument-level provenance contracts, the granularity \sys adopts; we did not run PACT and make no
quantitative claim against it. Table~\ref{tab:related} summarizes these designs.

\noindent\textbf{AuthGraph.} AuthGraph~\cite{authgraph2026} is the closest design: it checks the
sources of call parameters against an authorization specification derived from the user's request.
We did not run it. As described, AuthGraph authorizes a parameter through the \emph{source} it flows
from, whereas \sys compares the \emph{value} with the contract, or with the request text. If the
specification admits the company directory as a source of recipients, a source-based check accepts
\texttt{legal-team@company.com} and \sys rejects it unless the request writes that address. Whether AuthGraph's specifications
exclude such sources is untested.

\begin{table}[htbp]
\caption{Positioning, from each paper's own description and evaluation.}
\label{tab:related}
\centering\footnotesize
\setlength{\tabcolsep}{1.5pt}
\begin{tabular*}{0.6\linewidth}{@{\extracolsep{\fill}}lccccc}
\toprule
System & Gran. & Intent & Prov. & Adapt. & T-b-U\\\midrule
Spotlighting & prompt & -- & -- & -- & --\\
FIDES & planner & $\sim$ & \checkmark & -- & --\\
CaMeL & value & $\sim$ & \checkmark & -- & --\\
AttriGuard & tool & \checkmark & $\sim$ & -- & --\\
ClawGuard & tool & -- & $\sim$ & -- & --\\
Progent & tool & $\sim$ & -- & \checkmark & --\\
PACT & arg. & $\sim$ & \checkmark & -- & --\\
AuthGraph & tool, arg. & \checkmark & \checkmark & -- & $\sim$\\
\sys & arg. & \checkmark & \checkmark & $\sim$ & \checkmark\\
\bottomrule
\end{tabular*}

\vspace{2pt}{\footnotesize Gran.: granularity. Adapt., T-b-U: adaptive and trusted-but-unauthorized attacks evaluated in the paper. \checkmark: yes; $\sim$: partly; --: no or not reported. \sys's adaptive set was used during development. Our CaMeL\st re-implementation blocks all T-b-U attacks of Table~\ref{tab:disambig}.\par}
\end{table}

\section{\uppercase{Limitations}}

\noindent The guarantees of \sys depend on correct roles, intent, provenance labels, registry and
alias directory, and policy configuration. Table~\ref{tab:limitations} lists what it does not solve.

\begin{table}[htbp]
\caption{Limitations of \sys and of our study.}
\label{tab:limitations}
\centering\footnotesize
\setlength{\tabcolsep}{2pt}
\begin{tabular}{>{\raggedright\arraybackslash}p{0.30\linewidth}>{\raggedright\arraybackslash}p{0.66\linewidth}}
\toprule
Limitation & Consequence and current mitigation\\\midrule
One planner model & Live runs use \texttt{gpt-4o-mini} only; oracle-intent results do not depend on the planner\\
Synthetic main benchmark & One template generator; complemented by the T-b-U and AgentDojo-derived sets\\
Simplified baselines & Re-implementations (``-style''), not the authors' code; no ranking claim\\
Adaptive set seen in design & Results on it are optimistic; held-out campaign planned\\
Oracle intent and labels & Injection labels and intent are given; inference misses 25.9\% of authority roles; extraction not stressed\\
Registry and alias directory & Curated by hand; mismatches cause false blocks (5/20 on the replay)\\
Intent matching & Any substring of the request is accepted, even if an injected document contains it; undeclared arguments checked only for injection\\
Content and payload fields & Phishing text can reach an authorized recipient; semantic provenance planned\\
Role confusion & Attacker data hidden in a content field (4/8 succeed); content fields are not inspected\\
Human Confirm & Simulated as always correct; fatigue and social engineering not studied\\
Chained tool calls & Re-mediated by design, not evaluated here\\
Trusted computing base & Mediator or registry compromise fails open; logs hold raw values (pseudonymization optional)\\
\bottomrule
\end{tabular}
\end{table}

\section{\uppercase{Conclusions}}
\label{sec:conclusion}

\noindent Securing a tool-using agent requires checking which authority-bearing values it passes as
well as which tool it calls. On our benchmarks, comparing each such value with the user's intent blocks
trusted-but-unauthorized values that provenance accepts, and provenance with SQL checks covers most of
the other fields. This protection costs utility when endpoints cannot be traced or resolved (46.7\% benign
success without confirmation, 93.3\% with it), and \sys does not stop harmful content sent to an
authorized destination. Our results rest on oracle intent, one live planner and a synthetic main
benchmark. Next steps are live extraction with inferred roles end to end, a held-out adaptive
campaign, a full AgentDojo run with a live planner, and a study of human confirmation.

\section*{\uppercase{Acknowledgements}}

\begin{itemize}[leftmargin=*,nosep]
\item \textbf{Conflicts of Interest:} The authors have no conflicts of interest to declare.
\item \textbf{Financial and other Support:} This research received no specific grant from any funding agency.
\item \textbf{Author Contributions:} Y.~Ameur: conceptualization, methodology, supervision, writing
(review and editing); P.~Mendes: software, investigation, data curation, writing (original draft);
S.~Bouzefrane: supervision, writing (review).
\item \textbf{Data Sharing:} Code, benchmarks, policies and decision logs (MIT license):
\href{https://doi.org/10.5281/zenodo.24233094}{\texttt{doi.org/10.5281/zenodo.24233094}},
tag \texttt{icaart2027-camera-ready},
\href{https://github.com/pm-mendes/intentflow}{\texttt{github.com/pm-mendes/intentflow}}.
Controlled results can be recomputed from these logs or regenerated offline with the released scripts;
results that involve a hosted model are released as logs only.
\item \textbf{AI Tools:} The authors used Claude (Anthropic), a large language model assistant, to
implement, test and debug the evaluation pipeline, to run and diagnose experimental campaigns
(Sections~\ref{sec:eval}--\ref{sec:e2e}), to draft and revise the text of all sections, and to write
the TikZ code of Figure~\ref{fig:architecture}. No image-generation model was used. Every number,
table and figure traces to a script or a result file in the artifact; the authors designed the
protocols and checked every result and claim.
\end{itemize}

\bibliographystyle{apalike}
{\small\bibliography{references_v5}}

\end{document}